\def\year{2022}\relax
\documentclass[letterpaper]{article} % DO NOT CHANGE THIS
\usepackage{aaai21}  % DO NOT CHANGE THIS
\usepackage{times}  % DO NOT CHANGE THIS
\usepackage{helvet} % DO NOT CHANGE THIS
\usepackage{courier}  % DO NOT CHANGE THIS
\usepackage[hyphens]{url}  % DO NOT CHANGE THIS
\usepackage{graphicx} % DO NOT CHANGE THIS
\usepackage{natbib}  % DO NOT CHANGE THIS AND DO NOT ADD ANY OPTIONS TO IT
\usepackage{caption} % DO NOT CHANGE THIS AND DO NOT ADD ANY OPTIONS TO IT
\usepackage{xcolor}
\nocopyright
\usepackage{makecell}
\usepackage{amsmath,amssymb}
\usepackage[ruled,vlined]{algorithm2e}
\usepackage[pagewise]{lineno}
\newcommand*\rot{\rotatebox[origin=l]{90}}
\DeclareMathOperator*{\argmax}{\textrm{arg}\,\max}
\DeclareMathOperator*{\argmin}{\textrm{arg}\,\min}
\title{Robust Unpaired Single Image Super-Resolution of Faces}
\author{
    Saurabh Goswami, Rajagopalan A. N.\\
}
\affiliations{
    Indian Institute of Technology, Madras\\
    Chennai- 600036, India\\
    ee18s003@smail.iitm.ac.in, raju@ee.iitm.ac.in\\

}
\begin{document}

\maketitle

%%%%%%%%% ABSTRACT
\begin{abstract}
    We propose an adversarial attack for facial class-specific Single Image Super-Resolution (SISR) methods. Existing attacks, such as the Fast Gradient Sign Method (FGSM) or the Projected Gradient Descent (PGD) method, are either fast but ineffective, or effective but prohibitively slow on these networks. By closely inspecting the surface that the MSE loss, used to train such networks, traces under varying degradations, we were able to identify its parameterizable property. We leverage this property to propose an adverasrial attack that is able to locate the optimum degradation (effective) without needing multiple gradient-ascent steps (fast). Our experiments show that the proposed method is able to achieve a better speed vs effectiveness trade-off than the state-of-the-art adversarial attacks, such as FGSM and PGD, for the task of unpaired facial as well as class-specific SISR.
\end{abstract}
%-------------------------------------------------------------------------

%%%%%%%%% BODY TEXT
\section{Introduction}
Most real-world applications of SISR involve dealing with objects of a particular class (e.g. identifying faces in a speeding vehicle, spotting abnormalities in a medical image or monitoring the production line in a factory etc.). The capturing conditions are hardly ever ideal, and so the images often contain excessive blur, noise, illumination variation, and a host of different artefacts that can cause typical SISR networks employed in such situations to fail.\par
One way of making such networks robust to these degradations is to train them with examples that are accordingly degraded. However, compiling a dataset that covers such a wide range of diverse degradations is daunting to say the least. A much easier and more intuitive alternative is to train a conditional Generative Adversarial Network (GAN) that allows us to inflict a desired perturbation on an image as well as control its characteristics by tweaking a conditioning random vector. \cite{bulat,saurabh,maeda2020} follow this methodology to synthetically degrade clean images before using them to train robust SISR networks. Given that (a) such networks are vulnerable to these degradations and (b) they do not produce stable outputs under varying degradations, this is akin to a black-box attack on the candidate SISR networks. \par
To turn this into a white-box adversarial attack, we need an algorithm that, given an image and an initial value of the conditioning random vector, would locate that value, in the neighbourhood of the initial value, which causes the loss function of the SISR network to attain its local maxima. State-of-the-art adversarial attacks such as FGSM \cite{fgsm} and PGD \cite{pgd} fit this description. These methods ascend the gradient of a loss surface to find out what perturbation in the input would cause the maximum loss at the end of a network. To ensure preservation of semantic content, the perturbations are bounded and it is within these bounds that these attacks are optimal. However, learned degradation models are trained to always preserve semantic content irrespective of the value of conditioning random variables, making the norm unnecessary and rendering FGSM and PGD sub-optimal at best. Moreover, PGD, being an iterative procedure, takes a long time to converge to the desired maxima.\par
We observe that for Mean Square Error (MSE) loss function and normally distributed conditioning random vector, moving around on any two-dimensional (2D) plane in the space of this vector traces out, at the end of an unpaired facial SISR network, a surface that looks like the exponential of a second order polynomial and can be parameterized as such. Once we find these parameters, the maxima (if it exists) can be algebraically calculated without any gradient computation. This observation enables us to propose a novel white-box adversarial attack where, for every combination of a clean image and random normal vector: (a) we calculate the gradient of MSE loss at that random vector, (b) consider a triangle oriented in the direction of the computed gradient and sample loss values at multiple points on this triangle, (c) fit an exponential of a second order polynomial to this loss profile with weighted least squares and, (d) calculate the random vector corresponding to the worst degradation using the parameters of the estimated function. We do the above for every training iteration of the SISR network and feed the degraded image thus obtained to the network for training. During inference, we simply feed Low Resolution (LR) images to the trained SISR module to get the super-resolved outputs.\par
The main contributions of our work are as follows:
\begin{itemize}
    \item To the best of our knowledge, ours is the first work that considers the problem of adversarial robustness with degradations learned from real-world samples.
    \item We propose a novel white-box adversarial attack that is compatible with learned degradations and which leverages the parameterizable nature of the loss landscape of unpaired SISR networks.
    \item Even though our work mainly focuses on facial SISR, we perform extensive experiments to investigate the scope of our method as a useful adversarial attack even outside this specific scenario.
\end{itemize}

%------------------------------------------------------------------------
\section{Previous Works}
Real images are rife with degradations such as motion blur, defocus blur, random noise etc. Even though motion blur has been useful in detecting splicing forgery in images \cite{harnessingMotionBlur}, recovering the latent motion \cite{bringingAlive}, and defocus blur has been used to infer depth from a single image \cite{depthFromMotion}, in cases where the image is very small and contains  only a single class, blur greatly reduces the recognizability of the image. Image deblurring, in itself, is a very challenging inverse problem and most previous works assume the blur to be uniform. \cite{regionAdaptive} address it by using deformable convolution layers to adaptively change the size of receptive field to tackle non-uniform blur. However, it still leaves the task of SISR unattempted. Early works of Super-Resolution \cite{rangeMap,resolutionEnhancement,robustComputationally} used multiple shifted low-resolution images of the same scene to retrieve the latent high resolution image conditioned on the motion between the LR images. The performance of these algorithms, however, are highly dependent on the motion estimates. To address this, in \cite{motionFree}, a motion free super-resolution was attempted by analytically deriving the relation for the reconstruction of the superresolved image from its blurred and downsampled versions. Recent works \cite{srgan,esrgan,perceptionDistortion} though, use deep neural networks to directly learn the mapping from LR to HR domain.
\subsection{Robust Facial SISR}
It was first shown in \cite{urdgn} that facial images can be ultra-resolved using adversarial training. \cite{tdae} extended this approach for noisy facial images by training an autoencoder SISR network that would have an additional decoder block at the beginning for denoising an input image. During training, they used a random noise model that does not cover blur, compression artifacts or even signal dependent noise. To train a facial SISR network suitable for real-world application, the noise model must adequately represent the degrdations seen in real facial images. \cite{bulat} was the first work to take cognizance of this fact. They proposed a method whereby, in a two-step process, real degradation could be modeled with a Generative Adversarial Network (GAN) and an SISR network could be trained with LR images synthetically degraded with this model. During every epoch, differently degraded versions of the same images were shown to the SISR network by manipulating a random vector at the input of the degradation model. However, this method has two shortcomings: (a) it fails to maintain a steady output as the degradation in an image varies and, (b) it does not attempt to maximize the challenge for the SISR network by showing it degradations that it would find the most difficult to resolve. \cite{saurabh} addressed the first issue by fashioning their SISR network as an autoender the encodings of which were encouraged to stay consistent under varying degradations with the use of an explicit consistency constraint. This is an example of incorporating robustness through regularizing the SISR network to remain smooth (exhibit minimal variations) as the input transitions from clean to noisy. In our work, we address the second shortcoming by proposing an adversarial attack that is tailored for unpaired SISR network and which ensures that at every training iteration, our SISR network encounters only that degradation that it would find the most challenging to resolve. 
\subsection{Adversarial Robustness}
The outputs of deep neural networks often change drastically with even a minimal perturbation in the input. Adversarial robustness immunizes networks against this by exposing them to their worst-case scenario. These attacks (i) inflict an additive perturbation on an input before feeding it to the network, (ii) climb the loss surface to converge to the perturbation that correponds to a local maxima and, (iii) keep the norm of the perturbation constrained within a small value $\epsilon$ to preserve the semantic content of the input. In \cite{fgsm}, the authors proposed an attck that linearizes the loss surface of the target network and takes a single step of fixed size along its gradient to calculate the adversarial perturbation. Requiring only a single forward and backward propagation, it is very fast. \cite{pgd}, showing that linearization is not effective for larger $\epsilon$, finds this worst degrdation through multiple iterations, each comprising a forward and backward propagation, in an extremely slow process. \cite{trades} decomposes the robust error of a classification network into a natural classification error and a boundary error. They showed that a good trade-off between robustness and accuracy can be achieved by attacking the boundary error of a classification network.\par
Performing multiple gradient ascent steps through the network per training iteration is frustratingly time-consuming, the inherently content-preserving nature of learned degradation models renders $\epsilon$ unnecessary, and SISR is not a classification task. These make all the above methods unsuitable for unpaired SISR. Observing and leveraging the parameterizable nature of the MSE-loss surface of such networks, we propose a simple, intuitive and effective attack for unpaired facial SISR that needs neither an $\epsilon$ nor multiple gradient ascent steps.
%------------------------------------------------------------------------
\section{Proposed Method}
\subsection{Motivation}
\cite{advsr} showed that SISR networks perform poorly on unseen degradations. In order to train a robust facial SISR network for real-world application, we would therefore need a dataset with paired examples of degraded LR and clean HR images. There is currently no dataset that matches this description. The most widely used workaround, proposed by \cite{bulat} is to learn a diverse degradation model $(f_{deg})$ from a dataset of actual degraded images and training an SISR network $(f_{sr})$ with the following objective:
\begin{equation}
    \hat{\theta} = \argmin_{\theta} \mathbb{E}_{x}[\mathcal{L}_{\theta}(f_{sr}(f_{deg}(x,z)),x)]
\end{equation}
where, $z$ is a random vector that acts as a cue for the degradations. However \cite{pgd} shows that the robustness developed through this objective would only be suboptimal. The correct objective, as per their recommendation, is as follows:
\begin{equation}
    \hat{\theta} = \argmin_{\theta} \mathbb{E}_{x}[\argmax_{z}\mathcal{L}_{\theta}(f_{sr}(f_{deg}(x,z)),x)]
\end{equation}
The inner maximization term here tries to find the $z$ that would cause the maximum loss. State-of-art attacks like PGD \cite{pgd} and FGSM \cite{fgsm} are capable of perform this inner maximization. However, these are meant only for the following norm-bounded degradation model,
\begin{equation}
    f_{deg}(x,\delta) = x+\delta \quad \textrm{where $\|\delta\|_{\infty} \leq \epsilon$} 
\end{equation}
which merely adds an optimized additive random noise to an image, as opposed to contaminating it with real degradation. Moreover, FGSM stops being effective at large perturbations, and PGD is highly time-consuming.\par 
We needed a method that can handle general degradations $f_{deg}(x,z)$ and is fast and accurate. We inspected the MSE loss landscapes of SISR networks under learnt degradation models and observed that these surfaces can be approximated with exponential of second order polynomials. The worst degradation then becomes an algebraic function of the coefficients of these polynomials. 
\subsection{Observations}
The $f_{deg}$ modules, as used in \cite{bulat,saurabh}, degrade an input image based on a random normal vector $z \sim \mathcal{N}(0,I_n)$. We trained our $f_{deg}$ with the specifications prescribed in these papers and observed the following about the MSE loss of our unpaired SISR network:
\begin{enumerate}
    \item For facial SISR, the loss landscape computed on most triangles formed by 3 randomly sampled $z$ had only a single peak (trough) and a gradual fall-off. Approximating them with exponential of second order polynomials (with only $50$ points) took us very close to the maxima (minima). 
    \item To validate our observation, we fit $800$ component radial basis functions (RBFs) to the losses obtained for $8000$ random $z$ samples on a single image: 
    \begin{equation}
        l(z) = \sum_{i=1}^{800} a_{i} \ e^{-\frac{1}{2}(z-\mu_i)^T\Sigma_{i}^{-1}(z-\mu_i)}
    \end{equation}
    where $l(z)$ is the loss at $z$, and $a_{i},\mu_i,\Sigma_{i}$ are the co-efficient, mean and standard deviation of the i-th RBF. We diagonalized the estimated $\Sigma_{i}$ to find out the standard deviations of the RBFs. The smallest standard deviations were found to be greater than $2$. 
    \item Originating from $\mathcal{N}(0,I_n)$, the $L_2$ distance $d$ between two randomly sampled $z$, according to \cite{distribution} has the following cumulative distribution function (CDF):
    \begin{equation}
        \label{eq: disty}
        F(d;n) = 1 - \frac{\Gamma \left(\frac{n}{2}, \frac{d^2}{4}\right)}{\Gamma \left(\frac{n}{2}\right)} 
    \end{equation}
    where $\Gamma$ is the upper incomplete gamma function. For a large $n$, $F(x;n) \approx 1$ at $x=2$ indicating that most pairs of $z$ would be separated by distances less than $2$. Since the peaks of loss surfaces are wider than gaussian function with standard deviation of 2, this verifies our observation that most triangles formed by 3 randomly sampled $z$ would only contain one extrema.  
    \item This behavior persists across training epochs, making it a property that we can exploit.
\end{enumerate}
\subsection{Overview of the Method}
To train all the modules from scratch, we need two datasets: (i) a dataset $X$ with clean HR images, (ii) a dataset $Y$ containing degraded LR images. 
\subsubsection{Learning a Degradation Model}
\begin{itemize}
    \item Our $f_{deg}$ takes in an HR image $x \in X$ and an $n$ dimensional random normal vector $z \sim \mathcal{N}(O,I_{n})$ and produces a synthetically degraded LR image $\hat{y}$ with similar semantic content as $x$ and degradation similar to the samples in $Y$.
    \begin{equation}
    \label{eq:dm}
        \hat{y} = f_{deg}(x,z)
    \end{equation}
\end{itemize}
\subsubsection{Solving the Inner Maximization Problem}
\begin{itemize}
    \item For each $x$, we randomly sample $z_1$ and $z_2$, calculate the gradient of loss on $z_1$, and consider another point $z_3$ along the direction of this gradient. For simplicity, $z_2$ and $z_3$ are kept equidistant from $z_1$. This forms a triangle oriented in the direction where the loss starts increasing from $z_1$.
    \item We sample $n$ number of points ($z_i$) inside these triangles as a convex combination of $z_1, z_2$ and $z_3$ with co-efficients $(\alpha_i,\beta_i)$ as follows
    \begin{equation}
    z_i = \alpha_i  z_1 + \beta_i  z_2 + (1-\alpha_i-\beta_i)  z_3    
    \end{equation}
    We combine each $z_i$ with $x$, compute the SR outputs $f_{sr}(f_{deg}(x,z_i))$ and calculate the corresponding MSE losses ($l_i$).
    \item We fit the following function to the curated set of $(\alpha_i,\beta_i,l_i)$,
    \begin{equation}
    \label{eq:loss}
        l_i = \ e^{-(a\alpha_i^2+b\beta_i^2+c\alpha_i\beta_i+d\alpha_i+e\beta_i+f)}
    \end{equation}
    with weighted least square as $(A^TW^TWA)^{-1}A^TW^TWB$ where:
    
    %\resizebox{0.87\linewidth}{!}{
    \begin{equation}
    A = 
    \begin{bmatrix}
    \alpha_1^2 & \beta_1^2 & \alpha_1\beta_1 & \alpha_1 & \beta_1 & 1 \\
    \alpha_2^2 & \beta_2^2 & \alpha_2\beta_2 & \alpha_2 & \beta_2 & 1 \\
    \alpha_3^2 & \beta_3^2 & \alpha_3\beta_3 & \alpha_3 & \beta_3 & 1 \\
    \alpha_4^2 & \beta_4^2 & \alpha_4\beta_4 & \alpha_4 & \beta_4 & 1 \\
    \hdotsfor{6}\\
    \hdotsfor{6}\\
    \alpha_{n}^2 & \beta_{n}^2 & \alpha_{n}\beta_{n} & \alpha_{n} & \beta_{n} & 1 \\
    \end{bmatrix},
    B = 
    \begin{bmatrix}
    - \ln{l_1}\\
    - \ln{l_2}\\
    - \ln{l_3}\\
    - \ln{l_4}\\
    .\\
    .\\
    - \ln{l_{n}}\\
    \end{bmatrix}
    \end{equation}
    %}
    and
    \begin{equation}
        W = \textrm{diag}([w(l_1) \ w(l_2) \ w(l_3) \ w(l_4) \ ... \ w(l_{n}) )]) \\
    \end{equation}
    with 
    \begin{equation}
        w(l_i) = \frac{1}{\sqrt{2\pi}\sigma(\mathbf{l})}(\ \exp{(\frac{l_i-\max(\mathbf{l})}{\sigma(\mathbf{l})})} + \ \exp{(\frac{l_i-\min(\mathbf{l})}{\sigma(\mathbf{l})})})
    \end{equation}
    where $\mathbf{l}$ is the vector containing all the $l_i$s and $\sigma(\mathbf{l})$ is the standard deviation of $\mathbf{l}$. Since we are interested in accurately locating the local maxima (or minima), we put more weight on the extreme points with $W$.
    \item Next, we find the $(\alpha_*,\beta_*)$ values corresponding to the extrema using the following relation.
    \begin{equation}
        \begin{bmatrix}
        \alpha_*\\\beta_*\\
        \end{bmatrix}
        =
        \begin{bmatrix}
        2a & c\\
        c & 2b
        \end{bmatrix}^{-1}
        *
        \begin{bmatrix}
        -d\\-e
        \end{bmatrix}
    \end{equation}
    The derivation for the same can be found in the supplementary material.
    \item We use Eq. \ref{eq:loss} to calculate the corresponding loss $l_*$, append it to the list of $(\alpha_i,\beta_i,l_i)$ and calculate the $z_*$ corresponding to the maximum loss value in the list.
\end{itemize}
A batch implementation of this algorithm has been provided in the supplementary material.
\subsubsection{Training the Super-Resolution Network}
\begin{itemize}
    \item We degrade all the images in a minibatch using their corresponding $z_*$ and train $f_{sr}$ with a combination of MSE Loss and an Adversarial Loss to recover the clean HR image.
    \begin{equation}
        \label{eq:sr}
        \hat{x} = f_{sr}(f_{deg}(x,z_*))
    \end{equation}
\end{itemize}

\subsection{The Key Modules: $f_{deg}$ and $f_{sr}$}
In this subsection, we detail the architectures and loss functions used for our two key modules. Since we mainly focuses on formulating the attack, we adopt modules used in the previous works with only a few minor modifications.
\subsubsection{The Degradation Module $f_{deg}$}
\begin{itemize}
\item \textbf{Architecture:} As shown in Eq. \ref{eq:dm}, our degradation module takes a clean HR image $x$ with a random vector $z$ and produces $\hat{y}$ that appears to be degraded like the samples in $\mathcal{Y}$ and is semantically similar to $x$. As shown in Fig. \ref{fig:generators}a, the entire module consists of $12$ residual blocks, $4$ of them contain $2\times$ average pooling layers for downsampling and 2 of them have pixelshuffle layers for upsampling. In Fig. \ref{fig:generators}a, Res($n_i,n_o$) and Conv($n_i,n_o$) are residual and convolutional layers with $n_i$ input channels and $n_o$ output channels. The residual blocks are similar to the ones used in \cite{bulat}.
\begin{figure*}[t]
    \centering
    \includegraphics[width=\textwidth]{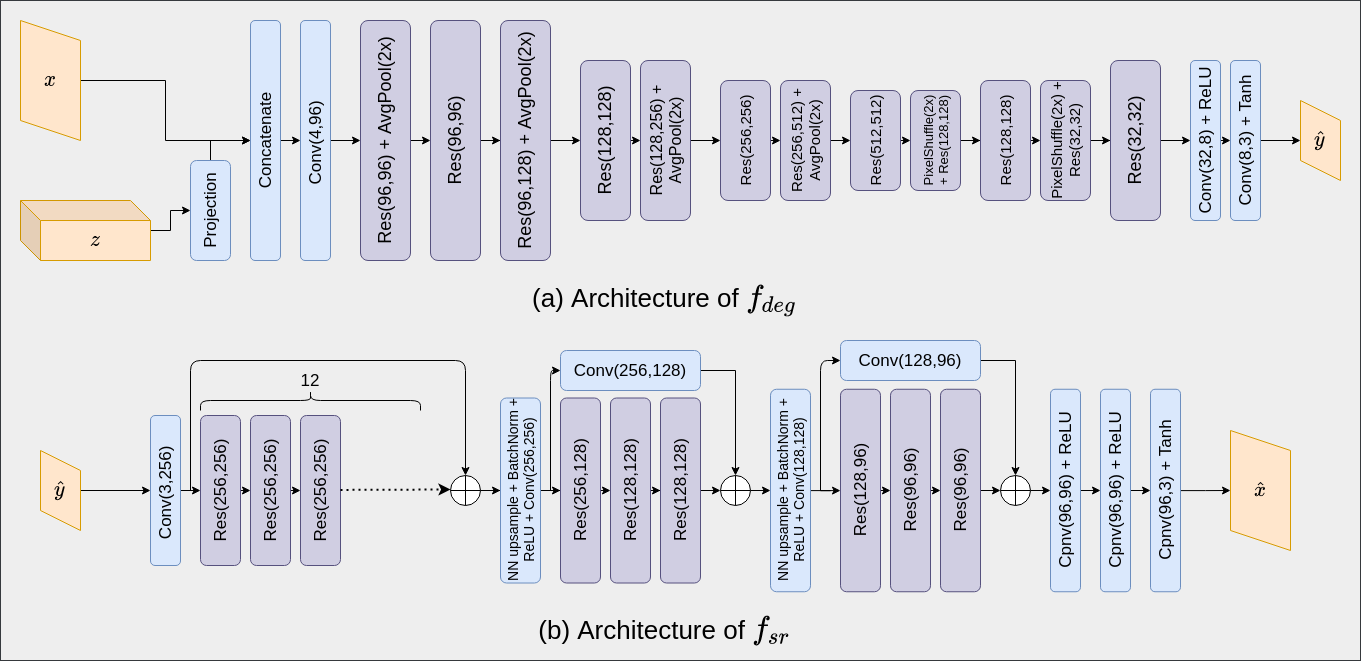}
    \caption{The key modules in our network.}
    \label{fig:generators}
\end{figure*}
\begin{figure*}[t]
\scriptsize
\centering
\begin{tabular}{cc}
    \begin{tabular}{cccccc}
         \rot{HR} & \includegraphics[width=0.05\textwidth]{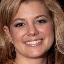} & \includegraphics[width=0.05\textwidth]{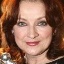} & \includegraphics[width=0.05\textwidth]{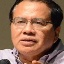} & \includegraphics[width=0.05\textwidth]{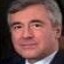} & \includegraphics[width=0.05\textwidth]{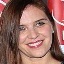}\\
         \rot{\makecell{ESR\\GAN}} & \includegraphics[width=0.05\textwidth]{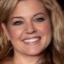} & \includegraphics[width=0.05\textwidth]{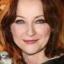} & \includegraphics[width=0.05\textwidth]{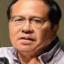} & \includegraphics[width=0.05\textwidth]{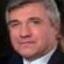} & \includegraphics[width=0.05\textwidth]{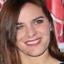}\\
         \rot{\makecell{Saurabh\\et al.}} & \includegraphics[width=0.05\textwidth]{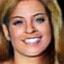} & \includegraphics[width=0.05\textwidth]{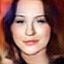} & \includegraphics[width=0.05\textwidth]{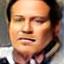} & \includegraphics[width=0.05\textwidth]{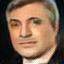} & \includegraphics[width=0.05\textwidth]{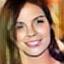}\\
         \rot{\makecell{Bulat \\et al.}} & \includegraphics[width=0.05\textwidth]{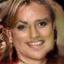} & \includegraphics[width=0.05\textwidth]{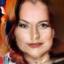} & \includegraphics[width=0.05\textwidth]{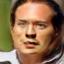} & \includegraphics[width=0.05\textwidth]{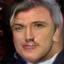} & \includegraphics[width=0.05\textwidth]{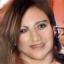}\\
         \rot{\makecell{Ours-\\FGSM}} & \includegraphics[width=0.05\textwidth]{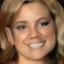} & \includegraphics[width=0.05\textwidth]{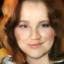} & \includegraphics[width=0.05\textwidth]{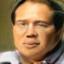} & \includegraphics[width=0.05\textwidth]{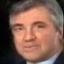} & \includegraphics[width=0.05\textwidth]{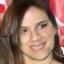}\\
         \rot{\makecell{Ours-\\PGD}} & \includegraphics[width=0.05\textwidth]{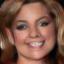} & \includegraphics[width=0.05\textwidth]{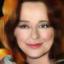} & \includegraphics[width=0.05\textwidth]{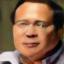} & \includegraphics[width=0.05\textwidth]{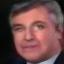} & \includegraphics[width=0.05\textwidth]{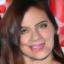}\\
         \rot{\makecell{Ours*}} & \includegraphics[width=0.05\textwidth]{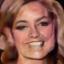} & \includegraphics[width=0.05\textwidth]{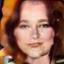} & \includegraphics[width=0.05\textwidth]{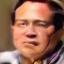} & \includegraphics[width=0.05\textwidth]{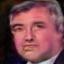} & \includegraphics[width=0.05\textwidth]{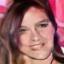}\\
         \multicolumn{6}{c}{(a) SISR results for clean LR images.}
    \end{tabular}
    &
    \begin{tabular}{cccccc}
        \rot{LR} & \includegraphics[width=0.05\textwidth,height=0.05\textwidth]{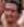} & \includegraphics[width=0.05\textwidth,height=0.05\textwidth]{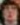} & \includegraphics[width=0.05\textwidth,height=0.05\textwidth]{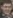} & \includegraphics[width=0.05\textwidth,height=0.05\textwidth]{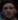} & \includegraphics[width=0.05\textwidth,height=0.05\textwidth]{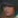}\\
         \rot{\makecell{ESR\\GAN}} & \includegraphics[width=0.05\textwidth]{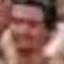} & \includegraphics[width=0.05\textwidth]{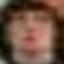} & \includegraphics[width=0.05\textwidth]{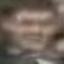} & \includegraphics[width=0.05\textwidth]{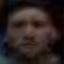} & \includegraphics[width=0.05\textwidth]{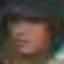}\\
         \rot{\makecell{Bulat\\et al.}} & \includegraphics[width=0.05\textwidth]{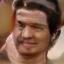} & \includegraphics[width=0.05\textwidth]{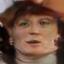} & \includegraphics[width=0.05\textwidth]{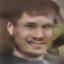} & \includegraphics[width=0.05\textwidth]{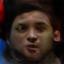} & \includegraphics[width=0.05\textwidth]{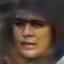}\\
         \rot{\makecell{Saurabh\\et al.}} & \includegraphics[width=0.05\textwidth]{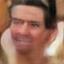} & \includegraphics[width=0.05\textwidth]{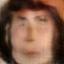} & \includegraphics[width=0.05\textwidth]{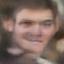} & \includegraphics[width=0.05\textwidth]{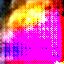} & \includegraphics[width=0.05\textwidth]{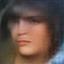}\\
         \rot{\makecell{Ours\\-FGSM}} & \includegraphics[width=0.05\textwidth]{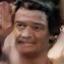} & \includegraphics[width=0.05\textwidth]{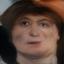} & \includegraphics[width=0.05\textwidth]{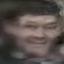} & \includegraphics[width=0.05\textwidth]{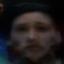} & \includegraphics[width=0.05\textwidth]{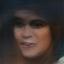}\\
         \rot{\makecell{Ours\\-PGD}} & \includegraphics[width=0.05\textwidth]{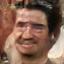} & \includegraphics[width=0.05\textwidth]{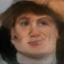} & \includegraphics[width=0.05\textwidth]{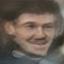} & \includegraphics[width=0.05\textwidth]{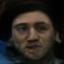} & \includegraphics[width=0.05\textwidth]{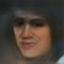}\\
         
         \rot{\makecell{Ours*}} & \includegraphics[width=0.05\textwidth]{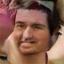} & \includegraphics[width=0.05\textwidth]{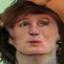} & \includegraphics[width=0.05\textwidth]{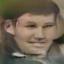} & \includegraphics[width=0.05\textwidth]{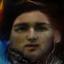} & \includegraphics[width=0.05\textwidth]{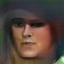}\\
         \multicolumn{6}{c}{(b) SISR results on degraded LR images.}
    \end{tabular}
    \end{tabular}
    \caption{\label{ressisr} Sample outputs for facial SISR.}
\end{figure*}
\item \textbf{Losses:} We impose an MSE loss between $\hat{y}$ and a bicubically downsampled version of $x$ (denoted here as $x_{\downarrow}$) to ensure the preservation of semantic contents.
\begin{equation}
    L_{deg}^{MSE} = \frac{1}{N}\|\hat{y}-x_\downarrow\|_2^2
\end{equation}
where $N$ is the dimensionality of $x$.\par
To ensure that the distribution of the outputs resembles $\mathcal{Y}$, we impose hinge GAN loss \cite{hingegan} (shown in Eq. \ref{eq:hingy1} and \ref{eq:hingy2}) on $\hat{y}$. We removed the sigmoid layer from the discriminator of SRGAN \cite{srgan} and imposed spectral normalization \cite{miyato2018spectral} on each layer to obtain our degradation discriminator $D_{d}$. The loss function of $D_{d}$ is as follows:
\begin{multline}
\label{eq:hingy1}
    L_{D_d} = \mathbb{E}_{y \in Y}[\max(0,1-D_{d}(y))] + \\
    \mathbb{E}_{x \in X}[\max(0,1+D_{d}(f_{deg}(x,z)))] 
\end{multline}
Likewise, the adversarial loss used for $f_{deg}$ is:
\begin{equation}
\label{eq:hingy2}
    L_{deg}^{Adv} = - \mathbb{E}_{x \in X}D_{d}(f_{deg}(x,z))
\end{equation}
Hence, the complete loss function for $f_{deg}$ is
\begin{equation}
    L_{deg} = \lambda_1 L_{deg}^{MSE} + \lambda_2 L_{deg}^{Adv}
\end{equation}
\end{itemize} 
\subsubsection{The Super-Resolution Module $f_{sr}$}
\begin{itemize}
    \item \textbf{Architecture:} We use the same network architecture as \cite{bulat} for designing $f_{sr}$. It has 3 residual units with 12, 3, 3 covolutional blocks respectively. The first 2 units have $2\times$ upscaling blocks at the end. To avoid checkerboard artifacts, we have used Nearest Neighbour interpolation for upscaling instead of transposed convolution. As shown in Eq. \ref{eq:sr}, $f_{sr}$ takes a synthetically degraded LR image supplied by the $f_{deg}$ and super-resolves it into clean HR image. Fig. \ref{fig:generators}b shows a diagram of this network.  
   \item \textbf{Losses:} We train this network with a combination of pixel-wise MSE loss ($L_{sr}^{MSE}$) and adversarial loss ($L_{sr}^{Adv}$) defined as follows:
   \begin{align}
       L_{sr}^{MSE} &= \|x-\hat{x}\|_2\\
       L_{sr}^{Adv} &= - \mathbb{E}_{x \in X}D_{s}(f_{sr}(f_{deg}(x,z)))
   \end{align}
where the discriminator $D_s$ has the same architecture as $D_d$ and is trained with the following loss function:
\begin{multline}
    L_{D_s} = \mathbb{E}_{x \in X}[\max(0,1-D_{d}(x))] \\+ \mathbb{E}_{x \in X}[\max(0,1+D_{s}(f_{sr}(f_{deg}(x,z_*))))]
\end{multline}
So,
\begin{equation}
    L_{sr} = \lambda_3 L_{sr}^{MSE} + \lambda_4 L_{sr}^{Adv}
\end{equation}
\end{itemize}
\section{Experiments}
Since a differentiable degradation module is the key component to carry out an adversarial attack, \cite{bulat} and \cite{saurabh} are the only two methods the comparisons with which would make sense in terms of validating the effectiveness of an adversarial training for unpaired SISR. However, we also include ESRGAN \cite{esrgan} in our comparison to show how faithful the outputs of our network are to the corresponding ground-truth images.\par
Due to SISR being a regression task, we can only compare our adversarial attack with FGSM \cite{fgsm} and PGD \cite{pgd}, since the more recent alternatives, such as TRADES \cite{trades} and YOPO \cite{yopo}, are tailored for classification tasks.
\subsection{Datasets}
Since our goal is to devise an adversarial attack for unpaired facial SISR networks, we focus primarily on faces. We randomly sample 153446 images from the Widerface \cite{widerface} dataset to compile the degraded LR dataset, namely LRFace dataset. 138446 of these images were used for training and 15000 for testing. To compile the clean dataset, i.e. HRFace dataset,  we combined the entire AFLW \cite{aflw} dataset with 60000 images from CelebAMask-HQ \cite{CelebAMask-HQ} dataset and 100000 images from VGGFace2 \cite{vggface2} dataset.\par
To see whether our attack would enhance the robustness of SISR networks in the larger class-specific as well as class-agnostic general unpaired SISR setting, we perform identical experiments on datasets of handwritten digits and natural scenes respectively. For handwritten digits, we took the original MNIST \cite{mnist} dataset as clean LR dataset, $4\times$ upscaled (using ESRGAN) counterpart as clean HR dataset, and the n-MNIST \cite{nmnist} dataset as degraded LR dataset. For SISR of natural scenes, we used the entire RealSR \cite{realsr} dataset. $80\%$ of all datasets were used for training, and $20\%$ for testing. 
\subsection{Implementation Details}
For all our experiments that involved implementing our method, we used $50$ points per sample to estimate the loss surface ($r=10,s=5$). We have performed all the training with a learning rate of $10^{-4}$ with Adam Optimizer ($\beta_1=0, \beta_2=0.9$) and a hyperparameter setting of $(\lambda_1=1, \lambda_2=0.05,\lambda_3=1,\lambda_4=0.05)$. For all experiments, we have kept the scale factor of SISR at 4. We keep the dimensionality of $z$ at $64$ and for every input image, we compute $50$ points on the corresponding triangles of $z$ to carry out the adversarial attack. For PGD, we use $20$ iterations with a step size of $0.01$. For both PGD and FGSM, $\epsilon=1$ has been used.
\subsection{Adversarial Attack}
This experiment tests the effectiveness of our method as an adversarial attack for SISR networks with learned degrdadation modules. We sample $1900$ clean HR images from HRFace and we paired them with $1900$ different $z$. We then locate their $z_*$ using FGSM \cite{fgsm}, PGD \cite{pgd} and our method. We calculate the average of the PSNRs and SSIMs entailed by all $z_*$s at the end of an undefended SISR network. Table \ref{tab:advattack} shows the results as well as the average time taken to perform each attack.\par
\begin{table}[h]
    \begin{center}
    \scriptsize
    \begin{tabular}{|c|c|c|c|}
    \hline
    \textbf{Dataset} & \textbf{FGSM} & \textbf{PGD} & Ours\\
    \hline
    \hline
    HRFace & \makecell{20.0384/\\0.6526} & \makecell{19.9394/\\0.6489} & \makecell{\textcolor{red}{19.3752}/\\\textcolor{red}{0.6264}} \\
    \hline
    MNIST & \makecell{15.6800/\\0.7992} & \makecell{15.5882/\\0.7966} & \makecell{\textcolor{red}{15.1429}/\\\textcolor{red}{0.7829}} \\
    \hline
    RealSR & \makecell{22.9481/\\0.6090} &  \makecell{\textcolor{red}{22.3563}/\\\textcolor{red}{0.5906}} & \makecell{22.8332/\\0.6061} \\
    \hline
    \hline
    \textbf{Time/iteration} & \textcolor{red}{1.18s} & 4.87s & 2.20s \\
    \hline
    \end{tabular}
    \end{center}
    \caption{\label{tab:advattack}Comparison of PSNR/SSIM and the average time taken per iteration for different attacks.}
\end{table}
Since adversarial attacks are meant to harm the performance of a target network, a lower PSNR/SSIM here indicates a more potent attack. On HRFace, FGSM proves to be the weakest as well as the quickest attack. Our method though, proves stronger and faster than PGD which is an iterative attack. This is because our method already knows how the parameters of a loss surface relates to its maxima, whereas PGD merely climbs the surface with only a limited number of steps and a fixed step-size. Using more and finer steps will make PGD even slower, and less steps may not be enough for reaching maxima. This shows that for unpaired facial SISR, our method is an optimal adversarial attack with a good speed vs effectiveness trade-off. Performing this experiment on MNIST \cite{mnist}, we got the same results as in the case of HRFace, showing that our attack is optimal for a broader class-specific unpaired SISR setting. On RealSR \cite{realsr}, our method was outperformed by PGD, suggesting that our hypothesis of single peak on most random triangles may not hold true for SISR of real scenes, something which was verified in Section \ref{sec:landy}.
\subsection{Facial Super-Resolution}
Here, we compare our adversarially robust facial SISR network with other facial SISR networks. Each network, in this experiment, is tested on two separate, unrelated datasets of $15,000$ images each: (a) a dataset of clean LR images and, (b) a dataset of real-degraded LR images. For clean LR images, we report PSNR/SSIM between the outputs and the ground-truth HR images. A higher PSNR/SSIM indicates that the network is good at preserving semantic contents of an input. For degraded LR images, lacking the corresponding ground-truth, we report the Fr\'echet Inception Distance (FID) \cite{fid} between the distribution of outputs and that of clean HR facial images. A lower FID indicates a more potent robustness. Since our work aims at enhancing the robustness of an unpaired SISR network, FID is our principal metric of interest. A PSNR/SSIM comparable with the previous methods only assures us that it does so without compromising the accuracy of reconstruction.\par
\begin{table}[h]
    \begin{center}
    \scriptsize
    \begin{tabular}{|c|c|c|}
        \hline
        \textbf{Method} & \textbf{PSNR/SSIM} & \textbf{FID}\\
         \hline
         \hline
         ESRGAN & \textcolor{red}{19.2533}/\textcolor{red}{0.8825} & 78.852\\
         \hline
         Saurabh et al. & 18.5587/0.6012  & 25.572\\
         \hline
         Bulat et al. & 16.9386/0.5012 & 23.158 \\
         \hline
         Ours-FGSM & 17.2348/0.4986 & 22.938 \\
         \hline
         Ours-PGD & 17.169/0.5127 & 15.583 \\
         \hline
         Ours* & 17.0609/0.5038 &  \textcolor{red}{9.454} \\
         \hline
    \end{tabular}
    \end{center}
    \caption{\label{tab:sisr}Performance comparison for facial SISR. PSNR/SSIM were computed on outputs of bicubically downsampled LR images, FID was computed on outputs of real degraded images.}
\end{table}
In Tab. \ref{tab:sisr}, we see that ESRGAN \cite{esrgan} achieves the highest PSNR/SSIM. This is expected since it was trained exclusively on clean (bicubucally-downsampled) LR facial images. However, since it never saw real degradations during training, it gives the highest FID on degraded images. \cite{bulat} was trained exclusively on real-degraded images and ends up learning a slightly erroneous SR mapping which fails to preserve the identity, pose and expression of an input face. This leads to a lower FID but a poorer PSNR/SSIM performance on clean images. Our training batches are $\frac{1}{3}$rd part clean LR image and $\frac{2}{3}$rd part degrdaded LR image. So, the last 3 methods  in Table \ref{tab:sisr}, that refer to our network with different attacks, perform better than \cite{bulat} in terms of both PSNR/SSIM and FID. Ours*, the variant of our network that uses our proposed attack, gives the lowest FID among all methods without compromising too heavily in PSNR/SSIM. FGSM leads to the highest FID amongst all attacks and hence, is the least effective. \par
We perform similar experiments on both MNIST \cite{mnist} and RealSR \cite{realsr} datasets. They are included in the supplementary material.

\section{Conclusion}
We propose a fast and simple novel adversarial attack for class-specific SISR ($4\times$) networks with learned degradation modules. By analyzing the MSE loss surface of these network, we discovered their easily parameterizable nature and this paved the way for an adversarial attack that is, as we establish through our experiments, simple, fast and effective. Using this adversarial attack, we were able to train a facial SISR network that is more robust than the previous state-of-the-art networks. In our future works, we intend to study the loss surfaces of such networks in more depth and see whether there is a way to optimize our method even further.

\end{document}